\documentclass[10pt,twocolumn,letterpaper]{article}

\usepackage[pagenumbers]{cvpr} % To force page numbers, e.g. for an arXiv version

\usepackage{graphicx}
\usepackage{amsmath}
\usepackage{amssymb}
\usepackage{booktabs}
\usepackage{multirow}
\usepackage{graphicx}
\usepackage{stfloats}
\usepackage[table,xcdraw]{xcolor}

\usepackage[pagebackref,breaklinks,colorlinks]{hyperref}

\usepackage[capitalize]{cleveref}
\crefname{section}{Sec.}{Secs.}
\Crefname{section}{Section}{Sections}
\Crefname{table}{Table}{Tables}
\crefname{table}{Tab.}{Tabs.}

\DeclareUnicodeCharacter{211D}{\ensuremath{R}}
\def\cvprPaperID{6630} % *** Enter the CVPR Paper ID here
\def\confName{CVPR}
\def\confYear{2023}

\def\authorBlock{
    Puhua Jiang\textsuperscript{1,}\textsuperscript{2}\quad
    Mingze Sun\textsuperscript{1} \quad
    Ruqi Huang\textsuperscript{1}\\
    
    \textsuperscript{1}Tsinghua Shenzhen International Graduate School \\
    \textsuperscript{2}Peng Cheng Laboratory \\    
    {\tt\small \{jph21, smz22\}@mails.tsinghua.edu.cn\quad ruqihuang@sz.tsinghua.edu.cn} 
}
\begin{document}

\title{Neural Intrinsic Embedding for Non-rigid Point Cloud Matching}

\author{\authorBlock}

\twocolumn[{
\renewcommand\twocolumn[1][]{#1}
\maketitle
\begin{center}
    \captionsetup{type=figure}
    \centerline{\includegraphics[scale=0.5]{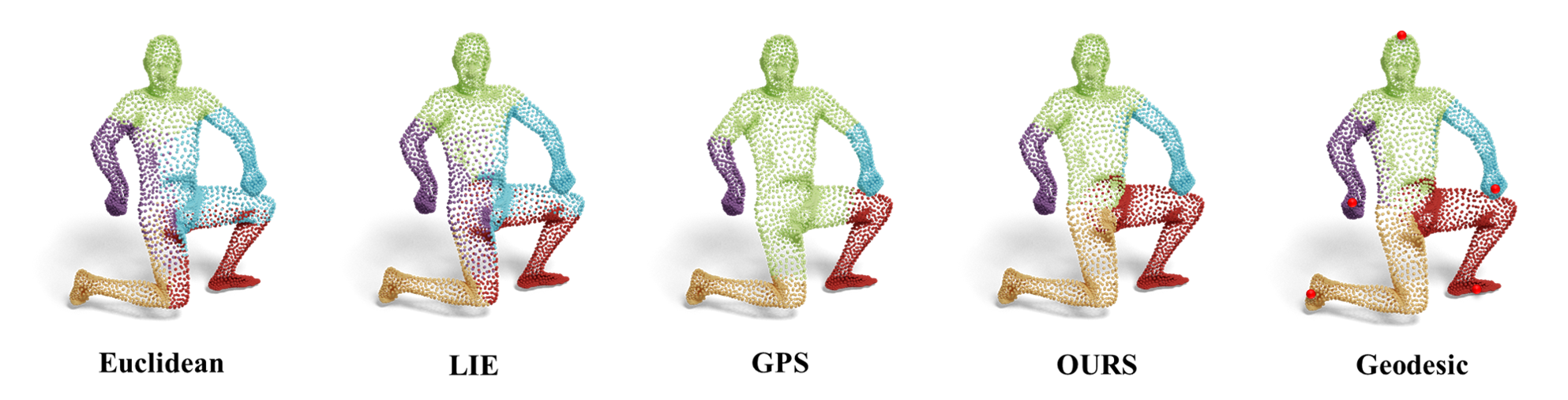}}

    \caption{Given a point cloud, we select 5 landmarks (see red points on the right-most one) and assign each of the rest points to the cluster represented by its nearest neighbor among the landmarks in the respective embedded space. We compare our method to Euclidean coordinates, LIE~\cite{lie}, GPS~\cite{lbo}. Our method takes in only point cloud and produces segmentation that is intrinsic geometry-aware. }
    \label{fig:teaser}

\end{center}
}]

%%%%%%%%% ABSTRACT
\begin{abstract}
As a primitive 3D data representation, point clouds are prevailing in 3D sensing, yet short of intrinsic structural information of the underlying objects. Such discrepancy poses great challenges on directly establishing correspondences between point clouds sampled from deformable shapes. In light of this, we propose Neural Intrinsic Embedding (NIE) to embed each vertex into a high-dimensional space in a way that respects the intrinsic structure. Based upon NIE, we further present a weakly-supervised learning framework for non-rigid point cloud registration. Unlike the prior works, we do not require expansive and sensitive off-line basis construction (e.g., eigen-decomposition of Laplacians), nor do we require ground-truth correspondence labels for supervision. We empirically show that our framework performs on par with or even better than the state-of-the-art baselines, which generally require more supervision and/or more structural geometric input.

\end{abstract}

%%%%%%%%% BODY TEXT

\section{Introduction}
\label{sec:intro}

Estimating correspondences between \emph{non-rigidly} aligned point clouds serves as a critical building block in many computer vision and graphics applications, including animation~\cite{jin2019fast, paravati2016point}, robotics~\cite{Gojcic_2019_CVPR, SANCHEZ2020}, autonomous driving~\cite{yue2018lidar, cui2021deep}, to name a few. 
In contrast to the well known rigid case, more sophisticated deformation models are in demand to characterize the non-rigid motions, for instance, articulation movements of human shapes. 

To address this challenge, extrinsic methods in principle approximate a complex global non-rigid deformation with a set of local rigid and/or affine transformations, e.g., point-wise affine transformation~\cite{li2018robust, yang2019global, wu2019global}, deformation graph~\cite{bozic2020deepdeform, li2020learning, bozic2020neural}, and patch-based deformation~\cite{xu2019unstructuredfusion, li2018articulatedfusion}. 
Being intuitive and straightforward, the extrinsic deformation models are in general redundant and lack global structures. 
%For instance, it is non-trivial to compose deformations using any of the above models. 
On the other hand, intrinsic methods~\cite{anguelov2004correlated, bronstein2006generalized, huang2008non, pai2021fast, moschella2021spectral, marin2021spectral} first transform extrinsic coordinates into an alternative representation, in which shape alignment is performed. 
For instance, the seminal functional maps framework~\cite{ovsjanikov2012functional} utilizes eigenbasis of the Laplace-Beltrami operator as spectral embeddings and turns non-rigid  3D shapes matching into rigid alignment of high-dimensional spectral embeddings, under the isometric deformation assumption. 
However, spectral embeddings are generally obtained by an inefficient, non-differentiable off-line eigen-decomposition of the Laplacian operator defined on shapes, either represented as polygonal meshes~\cite{cotangent} or point clouds~\cite{sharp2020laplacian}. 
Moreover, spectral embeddings are sensitive to various practical artifacts such as noise, partiality, disconnectedness, to name a few. 

To this end, we follow the isometric assumption and first propose a learning-based framework, Neural Intrinsic Embedding (NIE), to embed point clouds into a high-dimensional space. 
In particular, we expect our embedding to satisfy the following desiderata: (1) It is aware of the intrinsic geometry of the underlying surface; (2) It is computationally efficient; (3) It is robust to typical artifacts manifested in point clouds. 
Our key insight is that geodesics on a deformable surface, which are inherently related to Riemannian metric, contain rich information of the intrinsic geometry. 
Therefore NIE is trained such that the Euclidean distance between embeddings approximates the geodesic distance between the corresponding points on the underlying surface. 
In particular, considering the local tracing manner of geodesic computation, we choose DGCNN~\cite{wang2019dynamic} as our backbone, which efficiently gathers local features at different abstraction levels. 
We also carefully formulate a set of losses and design network modification to overcome practical learning issues including rank deficiency, and sensitivity to point sampling density. As a consequence, NIE manages to learn an intrinsic-aware embedding from merely unstructured point clouds. Fig.~\ref{fig:teaser} demonstrates that, we obtain the segmentation result the closest to the ground-truth based on geodesic distances. 

Furthermore, based on NIE, we propose a Neural Intrinsic Mapping (NIM) network, a weakly supervised learning framework for non-rigid point cloud matching. 
Though closely related to the Deep Functional Maps (DFM) frameworks, our method replaces the spectral embedding with the trained NIE, and further learns to extract the optimal features based on a self-supervised loss borrowed from~\cite{ginzburg2020cyclic}. 
In the end, we establish a pipeline for weakly supervised non-rigid point cloud matching, which only requires all the point clouds to be rigidly aligned and, for training point clouds, access to the geodesic distance matrices of them. 

Our overall pipeline is simple and geometrically informative. We conduct a set of experiments to demonstrate the effectiveness of our pipeline. In particular, we highlight that (1) our method performs on par with or even better than the competing baselines which generally require more supervision and/or more structural geometric input on near-isometric point cloud matching; (2) our method achieves sensible generalization performance, thanks to our tailored design to reduce bias of point sampling density; (3) our method is robust regarding several artifacts, including noise and various partiality.

\section{Related Work}\label{sec:related}
\noindent\textbf{Non-rigid point cloud matching}
This is a challenging task due to the complexity of modeling non-rigid deformations. Extrinsic methods~\cite{li2018robust, yang2019global, wu2019global, bozic2020deepdeform, li2020learning, bozic2020neural, xu2019unstructuredfusion, li2018articulatedfusion} approximate a complex global non-rigid deformation with a set of local rigid and/or affine transformations. 

On the other hand, intrinsic approaches, especially the spectral-based techniques, leverage the geometric information encoded in the eigenbasis of Laplacian operators, which lift the matching problem into a high-dimension space, where a family of isometric non-rigid deformations is well characterized. The structural benefits are attained at the cost of significantly larger search space for the optimal transformation. We conclude some typical spectral embeddings in the following.

\noindent\textbf{Geometric Embeddings }
Pioneered by the work~\cite{reuter2006laplace}, eigenbasis of the Laplace-Beltrami operator plays a dominant role in geometry processing for decades. Especially, several early approaches~\cite{lipman2010biharmonic, coifman2005geometric, lbo} attempt to establish connection between eigenbasis and surface geodesics, which encode essentially the intrinsic geometry. However, due to the computational burden and the noise-prone nature of high-frequency eigenfunctions, this line of works usually uses relatively low frequency eigenbasis, yielding only rough approximation in recovering geodesic distances. 

Related to this topic, there are also approaches directly optimizing for embeddings that best recover the underlying geodesics. For instance, MDS~\cite{torgerson1952multidimensional} is a classical dimension reduction method, which can achieve reasonably accurate embedding by minimizing certain stress. More recently, by exploiting the structural properties of the geodesics on the surface, GeodesicEmbedding~\cite{xia2021geodesicembedding} is proposed to build a hierarchical embedding, which in turn helps to reduce computing time of inferring geodesic distance on high-resolution meshes. While these approaches achieve relatively high recovery accuracy, we point out that they both require the \emph{ground-truth} geodesic distances as input, thus not suitable for our target. 

\noindent\textbf{(Deep) Functional Maps}
Another line of works that closely related to ours is the functional maps framework~\cite{ovsjanikov2012functional}. In the functional space, a correspondence can be represented by a small matrix encoded in a reduced eigenbasis and computed as the optimal transformation that aligns a given set of probe functions possibly with other regularization. 
Early works along this line take mostly an axiomatic approach~\cite{nogneng2017informative, huang2017adjoint,huang2014functional,kovnatsky2013coupled}, while in recent years a trend of integrating functional maps mechanism into a learning pipeline is attracting extensive attention~\cite{roufosse2019unsupervised, halimi2019unsupervised, eisenberger2021neuromorph, litany2017deep}
\\While most of the deep functional maps frameworks follow the utility of spectral embeddings and refine features upon some hand-crafted descriptors, e.g., HKS~\cite{sun2009concise}, WKS~\cite{WKS}, SHOT~\cite{tombari2010unique}. By leveraging full~\cite{donati2020deep} or weak~\cite{sharma2020weakly} supervision, networks are capable of extracting features directly from point clouds. Furthermore, exploration on how to establish embeddings to take over the spectral ones is also taken in~\cite{lie}, which again relies heavily on the supervision over shape correspondences.

\section{Background}\label{sec:bg}

For the sake of completeness, we briefly review the basic notions of functional map~\cite{ovsjanikov2012functional}, deep functional maps framework and the framework of Linear invariant embedding~\cite{lie} (LIE), which are closely related to our framework. 

\noindent\textbf{Functional Maps}
Functional maps~\cite{ovsjanikov2012functional} is an alternative representation of point-wise maps, which is formulated primarily upon eigenbasis of the Laplace-Beltrami operator. 
Given a pair of shapes $S_1, S_2$, one first computes the first $k$ eigenfunctions and store them as matrices $\Phi_i \in \mathbb{R}^{n_i \times k}, i = 1, 2.$
Now, given a point-wise map encoded as a permutation matrix $\Pi_{21}\in \mathbb{R}^{n_2\times n_1}$, the functional representation is 
\begin{equation}\label{eqn:convert}
	C_{12} = \Phi_2^{\dagger} \Pi_{21} \Phi_1 \in \mathbb{R}^{k\times k}, 
\end{equation}
where $\dagger$ denotes the Moore Penrose pseudo-inverse. Regarding the inverse conversion, one can compute via nearest neighbor search between the rows of $\Phi_2 C_{12}$ and that of $\Phi_1$.

One of the key properties of functional maps is that, by introducing the spectral embeddings, i.e., $\Phi_1, \Phi_2$, one can express \emph{global} map priors in simple algebraic forms in terms of $C_{12}$. For instance, area-preserving maps are supposed to correspond to orthogonal functional maps. In other words, one can add $\Vert C_{12}^T C_{12} - I\Vert_2$ as regularization to promote such property. 

\noindent\textbf{Deep Functional Maps} 
The above insight in turn gives rise to Deep Functional Maps (DFM) frameworks, which was first proposed in~\cite{litany2017deep}. 
In a nutshell, DFM is designed as a Siamese network, which aims to learn a universal feature extractor $\mathcal{G}: S_i \rightarrow G_i \in \mathbb{R}^{n_i\times d}$. Here $d$ is the number of features and $G_i$'s are assumed to be spectrally in correspondence. Therefore, one can formulate the following optimization problem:

\begin{equation}
	C_{12}=\underset{C \in \mathbb{R}^{k \times k}}{\arg \min }\left\|C_{12} \Phi_1^{\dagger} G_1-\Phi_2^{\dagger} G_2 \right\|_2+E_{\mbox{reg}}(C_{12})
\end{equation}

Equipped with the pre-computed spectral embeddings and some proper initial features (e.g., WKS~\cite{WKS}), one can optimize $\mathcal{G}$ over a set of training pairs, and output the optimal functional maps from the trained model, which can be converted to point-wise maps in the end. In fact, this is the basic design shared by several recent unsupervised DFM frameworks~\cite{roufosse2019unsupervised, halimi2019unsupervised, eisenberger2021neuromorph}

\noindent\textbf{Linearly Invariant Embedding~\cite{lie}}
It is evident that the key ingredient of functional maps representation is the spectral embeddings, which allow to encode point-wise maps into compact transformation matrices, but also to integrate and optimize map priors efficiently. 
LIE is the first work aiming to \emph{learn} a basis in place of spectral embedding. 

The key insight of LIE is that, given a collection of shapes and ground-truth correspondences among them, one can learn a basis generator that consumes a point cloud $X\in \mathbb{R}^{n_X\times 3}$ as input and return $k$-dimensional basis, i.e., 
\[\mathcal{F}(X) = \Phi_X\in \mathbb{R}^{n_X\times k}\]

Similar to Eqn.~\ref{eqn:convert}, given a ground-truth map $\Pi_{YX}$ from $Y$ to $X$, one can write the corresponding “functional map" as 		
\[C_{XY} = \Phi_Y^{\dagger} \Pi_{YX} \Phi_X.\]
Then LIE proposes to learn a basis generator $\mathcal{F}$, such that all the $C_{XY}$ with respect to the training pairs are \emph{orthogonal}. 
After $\mathcal{F}$ is learned, the authors further propose to learn a feature extractor $\mathcal{G}$ with the same training set and the correspondence labels, resulting in a DFM-like pipeline. 

Our framework share the same two-stage training strategy with LIE. 
However, we highlight that: (1) we pose no supervision on the correspondences across shapes; (2) our formulation is more geometrically informative; (3) unlike LIE, our method generalizes well even being trained within a small-scale dataset (see, e.g., Table~\ref{tab:ssf}).

\section{Method}
\label{sec:method}

In this section, we first formulate our Neural Intrinsic Embedding (NIE) network, and propose a weakly supervised matching network based on NIE, which we term as Neural Intrinsic Mapping (NIM) network. 
In general, for training our networks, we assume to be given a set of rigidly aligned point clouds and the corresponding dense geodesic matrices, with respect to the underlying surfaces. 

Note that, at inference time, both NIE and NIM require only point clouds approximately rigidly aligned with those in training, with no need of any further structural information, e.g., triangulation. 

\subsection{Neural Intrinsic Embedding}

We denote by $X_i\in \mathbb{R}^{n_i \times 3}$ a point cloud, and $d_S$ the geodesic distance function regarding the underlying surface, which can be discretized as a dense matrix recording all pairwise geodesic distances. 
Note that we do not assume the meshes to share the same number of vertices, nor the identical triangulation. 

We denote by $\mathcal{F}_{\Theta_B}$ the network generating our embedding, where $\Theta_B$ is the learnable parameters, and by $\Phi_i = \mathcal{F}_{\Theta_B}(X_i) \in \mathbb{R}^{n_i\times k}$, where $k$ is the dimension of our embedding. 

Considering a shape $S_i$, let $v_p, v_q$ be two vertices on it. Then our ultimate goal is such that 
\begin{equation}
	\Vert \Phi_i(p,:) - \Phi_i(q, :)  \Vert_2 = d_S(v_p, v_q), \forall v_p, v_q \in X_i
\end{equation}
where $d_S$ denotes the geodesic on the surface, and $\Phi_i(p, :)$ is the $p-$th row of embedding $\Phi_i$, i.e., the embedding of $v_p$. For a lighter notation, we denote by $d_E^i(v_p, v_q) = \Vert \Phi_i(p,:) - \Phi_i(q, :)\Vert_2.$

It seems then plausible to train a network with the following loss
\[L(\Theta_B) = \sum_i \sum_{(p, q)\in S_i \in [n_i]^2}\big| d_E^i(v_p, v_q) - d_S(v_p, v_q)\big|^2\] 

\noindent\textbf{Relative Geodesic Loss} However, the above naive loss, using absolute geodesic error, is prone to favoring long geodesic distance preservation within the embedding. This would in turn hamper the local distance preservation, due to the limited capacity of network and the finite embedding dimension. 
Thus, we instead use the loss penalizing the relative geodesic error: 
\begin{equation}\label{eqn:relgeo}
L_{\mbox{G}}(\Theta_B)=\sum_i \sum_{(p, q)\in S_i} \frac{\big| d_E^i(v_p, v_q)  - d_S(v_p, v_q)\big|^2}{d_S(v_p, v_q)^2},  
\end{equation}

\noindent\textbf{KL Loss} Furthermore, since preservation of local geometry is critical for obtaining fine-grained correspondences, we strengthen short distance recovery from a statistical point of view as follows. 
Given a vertex $v_p \in S_i$, we compute the two distances from it to all the other vertices $[d_S(v_p, v_1), d_S(v_p, v_2), \cdots, d_S(v_p, v_n)]$ and $[d_E^i(v_p, v_1), d_E^i(v_p, v_2), \cdots, d_E^i(v_p, v_n)]$. 
We then define a distribution by: 
\[P_S^p(v_q) = \frac{ \exp(-\alpha d_S(v_p, v_q))}{\sum_{q'} \exp(-\alpha d_S(v_p, v_{q'}))}, \forall v_q\in S_i\]
Similarly we can define another distribution $P_E^p$ with respect to the embedded distance $d_E^i$. 
Then we define a loss based on KL-divergence between distributions:
\begin{equation}\label{eqn:kl}
	L_{\mbox{KL}}(\Theta_B) = \sum_i \sum_p KL(P_E^p, P_S^p)
\end{equation}

\noindent\textbf{Bijectivity Loss} Training with the two losses above, we observe that the relative geodesic error of the network saturates at $k = 8$. 
Interestingly, this finding also agrees with~\cite{xia2021geodesicembedding}, where the authors find that their MDS-like embedding method also saturates at the same dimension. 
One consequence of this saturated performance is that further increasing embedding dimension leads to rank deficiency, which results in irreversible transforms with respect to the rank-deficient embeddings.

To address this issue, we take a self-supervised approach. Namely, we apply furthest point sampling to sample $2000$ vertices from the $X_i$, and then we let $X_i^a, X_i^b$ be the first and second $1000$ vertices from the sampled vertices. Note that by construction they are evenly distributed on the surface and well separated. 
We then compute the point wise maps $T_{ab}, T_{ba}$ between $X_i^a$ and $X_i^b$ via simply nearest neighborhood searching, as the two point sets are on the same surface. 
Finally, we write the point-wise into the form of permutation matrices $\Pi_{ab}, \Pi_{ba}$, and according to Eqn.~\ref{eqn:convert}, we have
\[C_{ab} = \Phi_i^{b\dagger} \Pi_{ba}\Phi_i^a, C_{ba} = {\Phi_i^a}^{\dagger} \Pi_{ab}\Phi_i^b.\]

Finally, We formulate the bijectivity loss as follows:
\begin{equation}
L_{\mbox{B}}(\Theta_B)=\sum_{a, b, i}\left\|C_{ab} C_{ba}-I\right\|_F^2+\left\|C_{ba} C_{ab}-I\right\|_F^2. 
\end{equation}

Putting every piece together, the total loss is written as: 
\begin{equation}\label{eqn:total}
L_{\mbox{total}} =  \lambda_{1} L_{\mbox{G}} + \lambda_{2} L_{\mbox{KL}} + \lambda_{3}L_{\mbox{B}}.
\end{equation}
where $ \lambda_{1}$, $ \lambda_{2}$ and $ \lambda_{3}$ are hyper-parameters. 

\begin{figure}[t]
\centerline{\includegraphics[scale=0.41]{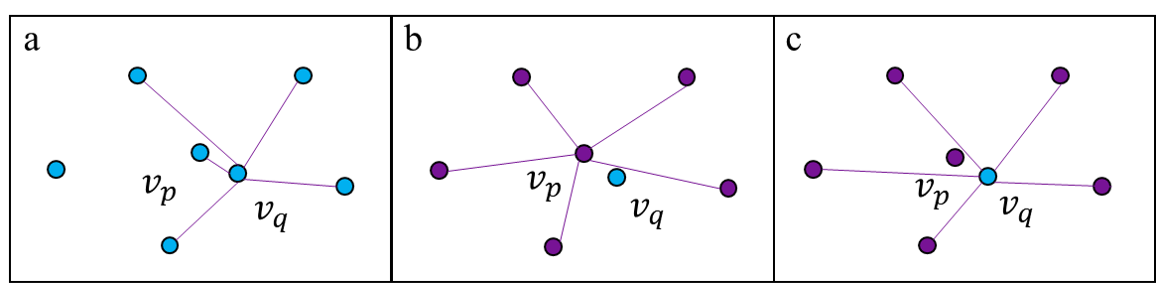}}
    \caption{Illustration of our aggregation method. (a)  the original aggregation method will include $v_p$ as the neighbor of $v_q$. (b) in our method, we find the $k-$NN of $v_p$ within $X_s$ . (c) we assign the neighbors of $v_p$ to $v_q$.}
    \vspace{-1em}
    \label{fig:edgec}
\end{figure}

\noindent\textbf{Alleviation of Sampling Density Bias:}
Apart from the aforementioned issues, we also encounter another problem hindering training -- point clouds may manifest varying sampling density across the underlying surface. 
Especially, the vanilla DGCNN implements local feature aggregation via $k-$nearest neighbor search, which is unaware of density distribution. This issue can significantly impact our generalization capacity, as each dataset owns its specific sampling pattern. To this end, we propose a simple yet effective modification on DGCNN as follows. 

Given a point cloud $X$, we first conduct furthest point sampling on $X$ to obtain a evenly distributed subset $X_s$. Now, given a point, $v_q$, instead of searching directly its $k-$NN within $X$, we first find its nearest neighbor, $v_p$, in $X_s$, and then assign the $k-$NN of $v_p$ within $X_s$ to $v_q$. 
The above description is well illustrated in Fig.~\ref{fig:edgec}, where $X_s$ are colored purple. 

In the end, we remark that sampling density bias is not a new issue --  several prior works~\cite{sharma2020weakly, lie, donati2020deep} that aim to learn feature/basis directly from non-rigid point clouds may have encountered the same problem. 
As a typical solution, the prior works also apply FPS sampling  to ensure a relative even distribution. In the case where spectral embeddings are available~\cite{sharma2020weakly, donati2020deep}, the authors simply leverage the fact that eigenbasis is insensitive to point distribution and estimate only functional maps. On the other hand, LIE~\cite{lie} circumvents this problem by heavily downsampling in both train and test point clouds (to $1k$ vertices), resulting in a dataset of low resolution. 
We highlight in Table~\ref{tab:ssf}, by utilizing our modified DGCNN, the generalization capacity of our method is largely enhanced.

\subsection{Neural Intrinsic Mapping}
\begin{figure}[b!]
\centerline{\includegraphics[scale=0.44]{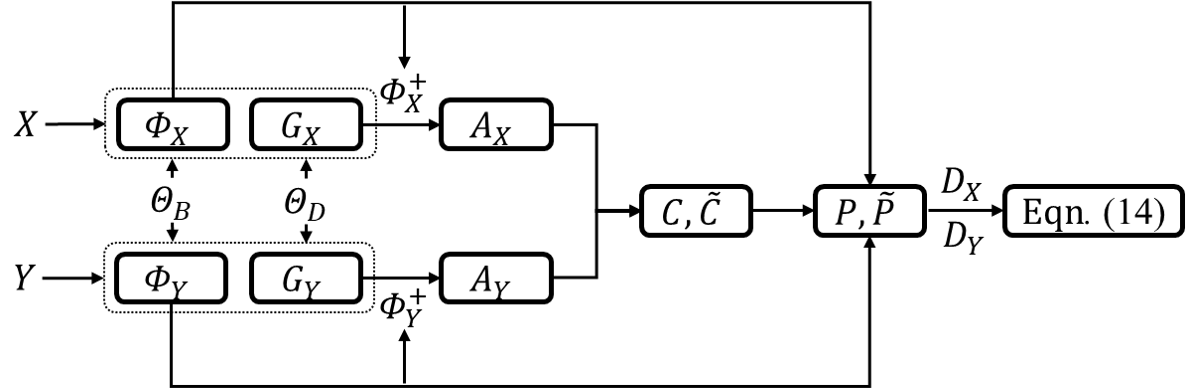}}
    \caption{Illustration of our pipeline. After computing the functional map $C$, we convert it into soft correspondence
map $P$ which is finally fed into the self-supervised loss.}
    \label{fig:pipeline}
\end{figure}

In this section, we formulate our NIM network. 
In essence, the network belongs to the family of the deep functional maps reviewed in Section~\ref{sec:bg}, though bears two main modifications as shown in Fig.~\ref{fig:pipeline}: (1) we replace the pre-computed eigenbasis with NIE proposed in Section~\ref{sec:niebasis} (denoted by $\Phi_X, \Phi_Y$ in the figure); (2) we remove the original structural losses on functional maps and instead use a self-supervised loss introduced in~\cite{ginzburg2020cyclic}, which is defined in terms of geodesic information to guide feature learning.  

In a nutshell, NIM learns to predict a set of optimal descriptors $G_{X} = \mathcal{G}_{\Theta_D}(X)$ and $G_{Y}=\mathcal{G}_{\Theta_D}(Y)$ from input point clouds $X$ and $Y$, here $\Theta_D$ is the set of learnable parameters. Once learned, the map from $Y$ to $X$ encoded in our NIE is given by:
\begin{equation}\label{eqn:fmapyx}
    C = A_X A_Y^{\dagger} = \left(\Phi_{X}^{\dagger} G_{X}\right)\left(\Phi_{Y}^{\dagger} G_{Y}\right)^{\dagger}
\end{equation}
Similarly, we have $\tilde{C}$ form $X$ to $Y$:
\begin{equation}
    \tilde{C} = A_Y A_X^{\dagger} = \left(\Phi_{Y}^{\dagger} G_{Y}\right)\left(\Phi_{X}^{\dagger} G_{X}\right)^{\dagger}
\end{equation}

Since we do not need any correspondence label, in order to make full use of the geodesic distance information, we convert the functional map $C$ into soft correspondence map and follow deep cyclic mapping~\cite{ginzburg2020cyclic} to design our unsupervised loss.
Given $C$, $\Phi_{X}$, $\Phi_{Y}$ the soft correspondence matrix mapping between input point clouds $X$ and $Y$ can be computed as:
\begin{equation}\label{eqn:alpha}
    P = \mbox{softmax} (- \alpha\Vert\Phi_{X}C - \Phi_{Y}\Vert_{2})
\end{equation}
where each entry $P_{ji}$ is the probability the $j-$th point in $X$ corresponds to the $i-$th point in $Y$, and $\alpha$ is a hyper-parameter controlling the entropy of the probability distribution. 

Similarly, we can compute the map in the opposite direction:
\begin{equation}
    \tilde{P} = \mbox{softmax} (- \alpha\Vert\Phi_{Y}\tilde{C} - \Phi_{X}\Vert_{2})
\end{equation}

Then the cyclic distortion~\cite{ginzburg2020cyclic} is 
\begin{equation}\label{eq:clcloss}
\begin{aligned}
L_{\text {cyclic }}(X, Y)=& \frac{1}{|X|^{2}}\left\|\left(D_{X}-(\tilde{P} P) D_{X}(\tilde{P} P)^{T}\right)\right\|_{F}^{2}+\\
& \frac{1}{|Y|^{2}}\left\|\left(D_{Y}-(P \tilde{P}) D_{Y}(P \tilde{P})^{T}\right)\right\|_{F}^{2},
\end{aligned}
% \end{multline} 
\end{equation}

where $D_{X}, D_Y$ are the geodesic distance matrix regarding $X$ and $Y$, respectively.

The above cyclic loss only encourages bijectivity of maps estimated from our NIM along different directions. As we always assume that the shapes of interest are near-isometric to each other, we take into consideration the following loss:

\begin{equation}\label{eq:isoloss}
\begin{aligned}
L _{\mbox{isometric}}( X , Y )=&\frac{1}{| X |^2}\left\|\left(D_{ X }-\tilde{P} D_Y \tilde{P}^T\right)\right\|_{ F }^2 +\\
& \frac{1}{| Y|^2}\left\|\left(D_{ Y }-P D_X P^T\right)\right\|_{ F }^2
\end{aligned}
\end{equation}

Thus the total loss for descriptor learning is:
\begin{equation}
L _{\mbox{desc}}( X , Y )= L _{\mbox{isometric}}( X , Y ) + L_{\mbox{cyclic}}( X , Y )
  \label{eq:important}
\end{equation}

\noindent\textbf{Map Inference via NIE and NIM}
Once we have trained the NIE and the NIM network, $\mathcal{F}_{\Theta_B}(\cdot), \mathcal{F}_{\Theta_D}(\cdot)$, we can estimate the correspondence between an pair of rigidly aligned point clouds $X$ and $Y$ as follows:
% \begin{itemize}
    (1) Compute neural intrinsic embeddings $\Phi_{X}, \Phi_{Y}$.
    (2) Compute the set of learned features, $G_{X}, G_{Y}$. 
    (3) Compute $C_{YX}$ according to Eqn.~\ref{eqn:fmapyx}.
    (4) Compute the point-wise correspondences as described in Section~\ref{sec:bg}.
% \end{itemize}

\section{Implementation}
\label{sec:formatting}

We implemented our pipeline in PyTorch~\cite{paszke2019pytorch} by adapting implementation of DGCNN~\cite{wang2019dynamic} released by the authors. Our network contains three EdgeConv layers mapping the input dimension from 3 to 64 and then to 512, followed by three convolutions layers reducing the dimension from 512 to output. For the basis generator network, we train it with a batch size of 3 for 600 epochs. We use a cosine annealing schedule with an Adam optimizer in between a maximum learning rate of $0.002$ and a minimum learning rate of $0.0002$. During training, we randomly sample $4995$ points from each shape. For the descriptor generator network, we use a batch size of 4, again with a cosine annealing schedule with an Adam optimizer in between a maximum learning rate of $0.002$, and a minimum learning rate of $0.001$. We use the same backbone for both basis and descriptor generator network, only the output feature dimension differs. We always train  with embedding dimension of $20$. For more details, please refer to the appendix.

\section{Experimental Results}
\label{sec:exp}

In this section, we demonstrate a set of experiments, comprised of three main parts as follows. 
First of all, in Section~\ref{sec:niebasis}, we evaluate our learned embeddings and provide ablation studies to justify our proposed design. 
Secondly, in Section~\ref{sec:matching}, we demonstrate the matching results of our proposed NIM network and compare it to several competitive baselines. 
Finally, in Section~\ref{sec:robust}, we demonstrate the robustness of our NIE and NIM network with respect to artifacts including noise and various partialities. 
We report all matching results in terms of mean \textbf{geodesic} error on shapes normalized to the unit area, even in the case that only point clouds are fed in inference time. \\

\noindent\textbf{Datasets} Before reporting our results, we provide details on the involved datasets:
\textbf{FAUST\_r:} The remeshed version~\cite{ren2018continuous} of FAUST dataset\cite{FAUST} contains 100 human shapes. We split the shapes as 80/20 for training and test. 
\textbf{SCAPE\_r:} The remeshed version~\cite{ren2018continuous} of SCAPE dataset\cite{anguelov2005scape} contains 71 human shapes. We split the shapes into 51/20 for training and test. 
\textbf{SURREAL\_r:} We randomly sampled 120 human shapes from SURREAL dataset~\cite{varol2017learning}, and perform remeshing so that each shape has around 5000 points. We split the shapes into 100/20 for training and test. 

\subsection{Embedding Evaluation}\label{sec:niebasis}

\noindent\textbf{Embedding Quality} \quad
We compare our NIE with several embeddings including the Euclidean coordinates (properly centered and normalized), MDS~\cite{torgerson1952multidimensional}, eigenbasis of the Laplace-Beltrami operator~\cite{lbo} defined on meshes, eigenbasis of the Laplacian operator~\cite{pclbo} defined on point clouds, and LIE~\cite{lie}. 
For a fair comparison, we set all embeddings to be of dimension $20$, with an exception of Euclidean coordinates.  
Regarding LIE and our method, we train the basis generator network on the $51$ training shapes from SCAPE\_r dataset, and evaluate all the basis, either constructed or learned, on the rest $20$ test shapes. 

% \ruqi{complete the reference.}

We evaluate all the embeddings via two metrics proposed before: (1) the relative geodesic error (Eqn.~\ref{eqn:relgeo}); (2) the following metric, termed as \emph{OPT}, introduced in~\cite{lie}:
Given a pair of point clouds $X, Y$, together with the ground-truth correspondence $\Pi_{YX}$, we first use Eqn.~\ref{eqn:convert} to encode the correspondence into a matrix regarding an embedding, then we recover the point-wise map from the matrix representation, and evaluate the geodesic error of the recovered map w.r.t the ground-truth.

\begin{table}[t!]
\centering
\begin{tabular}{ccc}
\hline
Method & OPT & Geo. Err.\\ \hline

Euclidean                  & 14.0                 & 19.5                     \\
MDS 20~\cite{torgerson1952multidimensional}                  & 3.3                & 12.0                       \\
LBO basis 20~\cite{lbo}             & 3.7                & 1271.1         \\
PCD LBO basis 20~\cite{pclbo}          & 3.8                & 1261.9        \\ \hline
LIE~\cite{lie}                     & 3.6                & 1543.1        \\
Ours  & \textbf{3.1}       & \textbf{9.5}                     \\ \hline
\end{tabular}
\caption{Comparative results of OPT (×100) and relative geodesic error (x100) of the different methods on basis.}
\label{tab:table1}
\end{table}

As shown in Table~\ref{tab:table1}, it is indeed expected that our method performs the best regarding the first metric, since we train our network using exactly the same loss. While there is no related constraint on LBO, PC-LBO and LIE, leading to significant relative geodesic errors. 
It is worth noting, though, our method outperforms MDS as well, which takes as input the ground-truth geodesic matrices. 
This is because MDS regresses embeddings with respect to the \emph{absolute} geodesic error, which naturally favors long distance preservation. 
And interestingly, in terms of \emph{OPT}, MDS20 is also outperformed by our method, suggesting the rationality of training with relative geodesic error.  

On the other hand, it is remarkable that our method performs best in \emph{OPT}. 
Especially, LIE enforces the encoded ground-truth maps to be orthogonal during training, which introduces strong structural prior on the \emph{OPT} metric, while our pipeline is trained without any supervision on correspondences across shapes. 

\begin{table}[t!]
\centering
\begin{tabular}{lccc}
\hline
Method & OPT & Geo. Err. & Mat. Err.  \\ \hline

 $L_{\mbox{G}}$               & 4.4                & \textbf{8.8}        &13.2       \\
$L_{\mbox{G}}+L_{\mbox{B}}$             & 3.5                & 12.4       & 11.8               \\
$L_{\mbox{G}}+L_{\mbox{B}}+L_{\mbox{KL}}$         & 3.3                & 10.6        & 11.5                \\
\mbox{Full model with sample}  & \textbf{3.1}       & 9.5        & \textbf{11.0}            \\ \hline
\end{tabular}
\caption{Ablation study of training loss on OPT (×100), relative geodesic error (x100), and the final matching error (x100). }\vspace{-2em}
\label{tab:table2}
\end{table}

\noindent\textbf{Ablation on NIE Design} \quad
In Table~\ref{tab:table2} we report ablation studies on the training loss terms and our modified DGCNN. 
When only the relative geodesic loss $L_{\mbox{G}}$ is used, though we can get the lowest error, NIE suffers from a rank deficiency problem, which in turn leads to the worst \emph{OPT} score. Adding the bijectivity loss $L_{\mbox{B}}$ effectively retains full rank and improves the \emph{OPT} score by $20\%$. Combining the KL loss $L_{\mbox{KL}}$, we further improve the OPT score as well as the relative geodesic error. Finally, integrated with our modified version of DGCNN, our full model performs the best in the ablation study. We also ablate the effect of losses on the final mapping accuracy. It is evident that each loss contributes to the final performance.

\subsection{Near-isometric point cloud matching}\label{sec:matching}

\textbf{Baselines} \quad We compare our method with a set of baselines, which are categorized depending on if mesh information is required during \emph{inference time}: (1) BCICP~\cite{ren2018continuous}, SURFMNet~\cite{roufosse2019unsupervised}, UnsupFMNet~\cite{halimi2019unsupervised}, NeuroMorph~\cite{eisenberger2021neuromorph}, FMNet~\cite{litany2017deep}, WSupFMNet~\cite{sharma2020weakly} in which meshes are required for computing eigenbasis; (2) 3D-CODED~\cite{groueix20183d}, CorrNet-3D~\cite{zeng2021corrnet3d}, LIE~\cite{lie}, on the other hand, can directly predict point-wise maps based on point clouds as test input. The used supervision is indicated next to each method in the table: \textbf{U}nsupervised, \textbf{S}upervised, \textbf{W}eakly-supervised.

\begin{table}[t!]

\begin{tabular}{c|cccc}
\hline
Method &  F & S & F on S & S on F \\
\hline
BCICP~\cite{ren2018continuous}                   & 15.             & 16.               & \textbackslash{}        & \textbackslash{}        \\
SURFMNet(U)~\cite{roufosse2019unsupervised}                                 & 15.               & 12.               & 32.                    & 32.                    \\
UnsupFMNet(U)~\cite{halimi2019unsupervised}          & 10.                & 16.               & 29.                    & 22.                    \\
NeuroMorph(U)~\cite{eisenberger2021neuromorph}           & 8.5              & 30.                & 29.                    & 18.                    \\
FMNet(S)~\cite{litany2017deep}                & 11.               & 12.               & 30.                     & 33.                    \\
WSupFMNet(W)~\cite{sharma2020weakly}            & \textbf{3.3}              & \textbf{7.3}     & \textbf{12.}           & \textbf{6.2}          \\ \hline
3D-CODED(S)~\cite{groueix20183d}             & \textbf{2.5}     & 31.               & 31.                    & 33.                    \\
CorrNet-3D(U)~\cite{zeng2021corrnet3d}           &  63.               & 58.               & 58.                    & 63.                    \\
LIE(S)~\cite{lie}                  & 3.6     & 12.               & 19.                    & 12.                    \\
Ours(W)                 & 5.5              & \textbf{11.}      & \textbf{15.}           & \textbf{8.7}          \\ \hline
\end{tabular}

\caption{Comparative results of mean geodesic errors (×100) of the different methods on Near-isometric point cloud matching. The \textbf{best} results are highlighted separately for methods with mesh and without mesh.}
\label{tab:table3}
\end{table}

First, we train models on FAUST\_r and SCAPE\_r datasets respectively. In particular, we train our NIE and NIM network both with ground-truth geodesic information computed on the meshes from the training set. 
In Table~\ref{tab:table3}, we report the normal matching errors as well as generalized matching errors. For instance, the column \textbf{F on S} reads that training on FAUST\_r but test on SCAPE\_r. 
In Table~\ref{tab:table3}, the best score from each category are highlighted in bold. Our method performs the best in $3$ out of $4$ terms among the competing methods of the same category. 
Indeed, our score is also the second best of \emph{all} methods in the table with respect to the $3$ terms, only being outperformed by WSupFMNet~\cite{sharma2020weakly} by a reasonable margin, given the fact that the latter uses eigenbasis of the Laplace-Beltrami operator.

\begin{table}[]
\centering
\begin{tabular}{ccc}
\hline
Method             & S             & F            \\ \hline
CorrNet-3D~\cite{zeng2021corrnet3d}         & 52.          & 54.         \\
LIE~\cite{lie}                & 20.          & 15.         \\
Ours               & \textbf{10.} & \textbf{6.5} \\ \hline
CorrNet-3D Noise & 58.          & 62.         \\
LIE Noise          & 20.          & 15.         \\
Ours Noise       & \textbf{11.} & \textbf{7.2} \\ \hline
\end{tabular}
\caption{Mean geodesic errors (×100) when trained on Surreal and tested on re-meshed Faust and Scape.}
\label{tab:ssf}
\end{table}

\begin{figure}[ht]
\centerline{\includegraphics[scale=0.26]{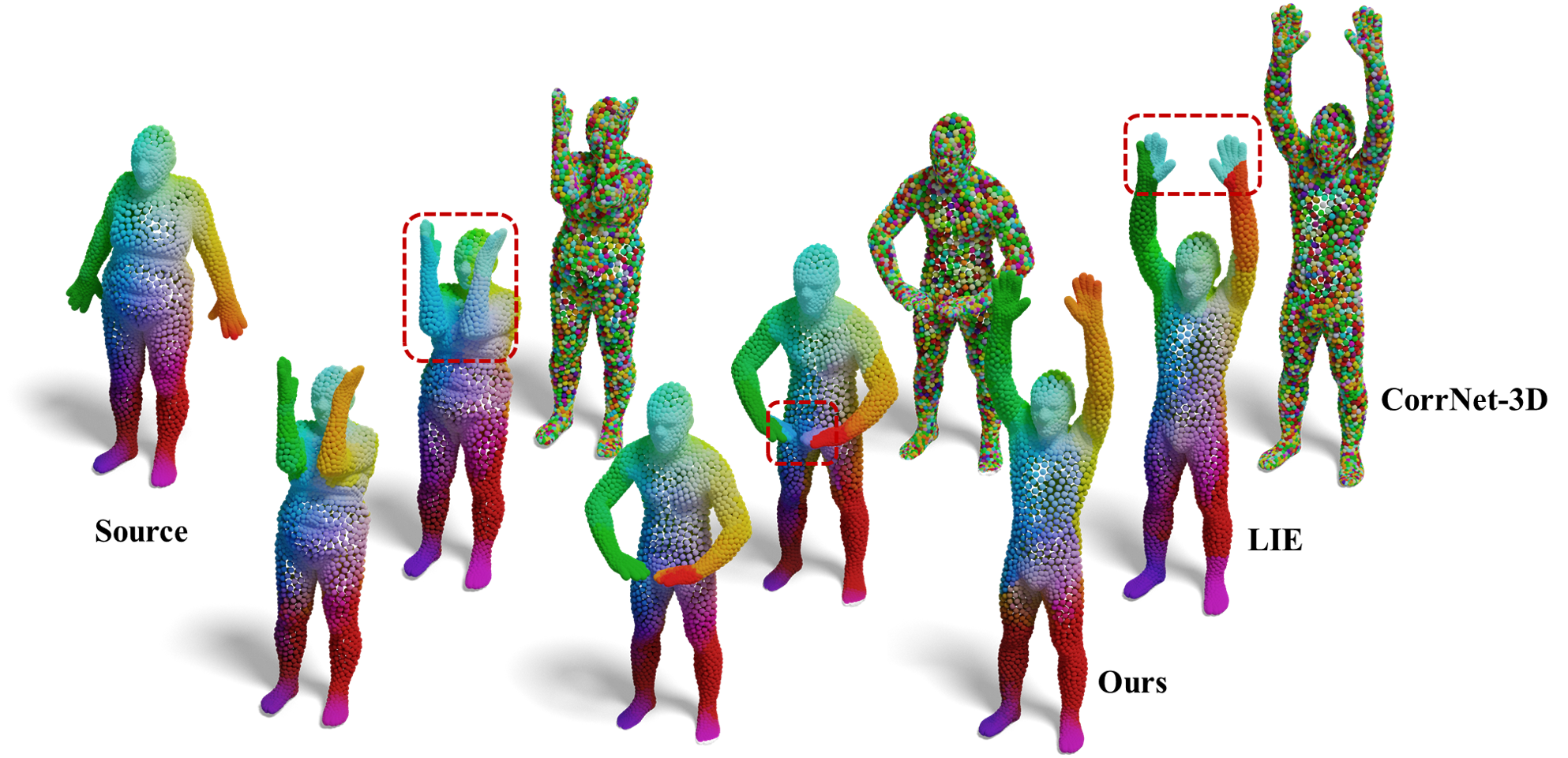}}
    \caption{Qualitative results of noise-free examples from FAUST\_r of Table \ref{tab:ssf}. CorrNet-3D fails in all three examples. LIE has obvious mismatches around the hands. While our method produces high-quality maps. }
    \label{fig:3nonoise}
\end{figure}

\begin{figure}[ht]
\centerline{\includegraphics[scale=0.26]{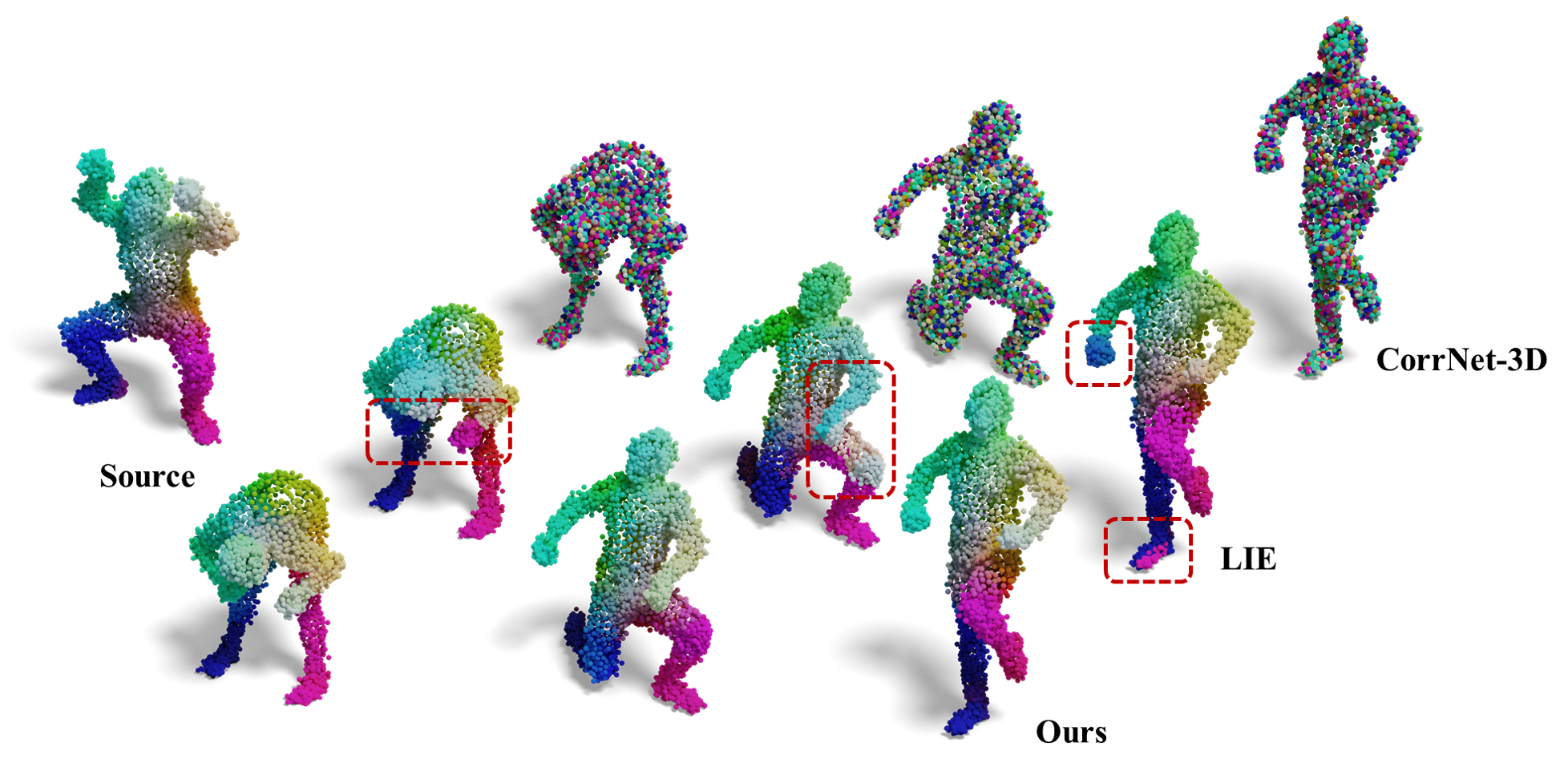}}
    \caption{Qualitative results of noisy examples from SCAPE\_r of Table \ref{tab:ssf}. The baseline methods suffer from the noise while our method still predicts reasonable maps.}\vspace{-1em}
    \label{fig:3noise}
\end{figure}
We report further the generalization capacity in Table~\ref{tab:ssf}. In this case, we train our NIE and NIM network, as well as the baseline methods, on SURREAL\_r, and then use the trained models to infer test shapes of SCAPE\_r and FAUST\_r. 
In this case, we mainly compare CorrNet-3D~\cite{zeng2021corrnet3d} and LIE~\cite{lie}. 
It is evident from the top half of Table~\ref{tab:ssf} that our method generalizes the best, with $50\%$ and $56.7\%$ matching error reduction upon LIE. 
We also provide qualitative illustrations on the computed maps from different approaches in Fig.~\ref{fig:3nonoise}.

Finally, we demonstrate that, given a trained NIE, one can even train a NIM network on a different training set, where geodesic information is absent. 
More specifically, we first train the NIE module on SURREAL\_r dataset. Then given a set of point clouds from other dataset, e.g., the training set of FAUST\_r, we can use the trained NIE to embed the unseen point clouds, and to approximate the geodesic distances with Euclidean distances among the embeddings. 
In the end, we train NIM with the point clouds from FAUST\_r and the respective approximated geodesics. 

\begin{figure}[t!]
\centerline{\includegraphics[width=7cm,height=5cm]{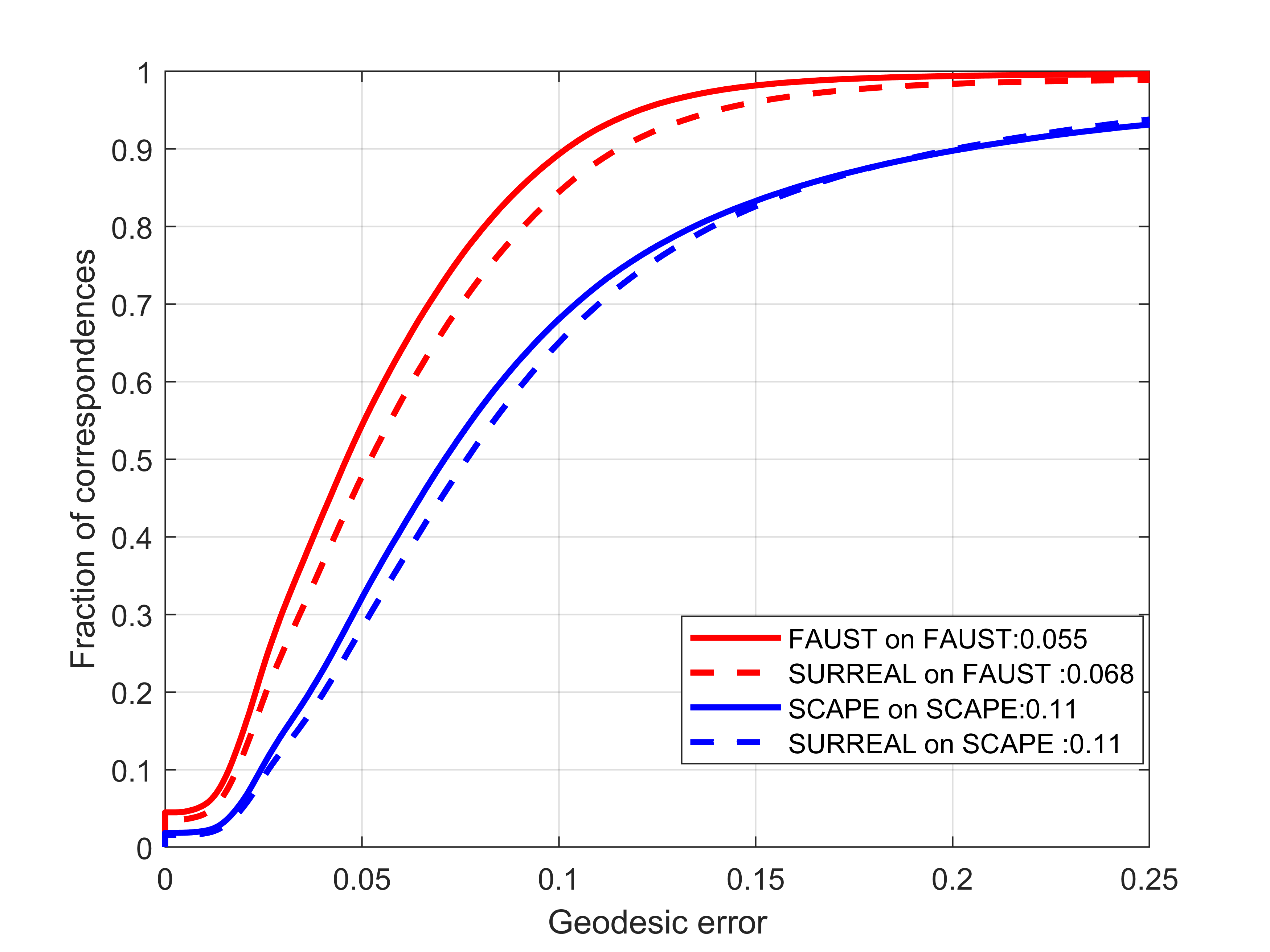}}
    \caption{Compare between directly inferring on datasets and fine-tuning on datasets. Our method achieves decent performance compared to the models trained with full information.}
%    \vspace{-2em}
    \label{fig:ft}
\end{figure}

Fig.~\ref{fig:ft} shows the results on the above learning protocol. As a strong baseline, we train two NIM's on FAUST\_r and on SCAPE\_r, which exploit the full information from the respective dataset. As shown in Fig.~\ref{fig:ft}, our method, without any ground-truth geodesic information from the dataset of interest, achieves decent performance even compared to the models trained with full information.

\subsection{Robustness}\label{sec:robust}

In this section, we show that our NIM network is robust with respect to typical artifacts including noise, various partiality, and even disconnectedness. 
We start our experiments following the setting presented in Table~\ref{tab:ssf}, however this time we perturb the input point clouds by Gaussian noise. As shown in the bottom half of Table~\ref{tab:ssf}, our accuracy still significantly outperforms the competing baselines by a large margin. We also provide qualitative evaluation in Fig.~\ref{fig:3noise}. 

\begin{figure}
\centering
\centerline{\includegraphics[scale=0.4]{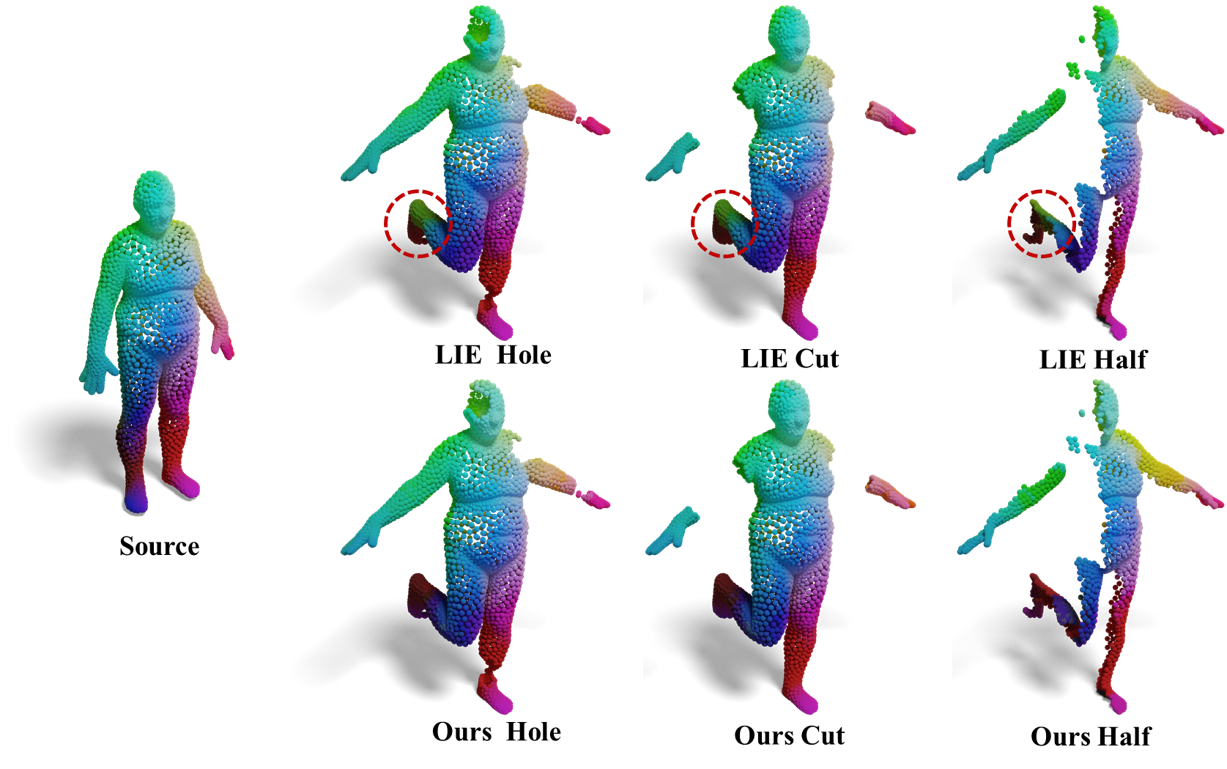}}
    % \vspace{-}
    \caption{Qualitative examples of partial point cloud matching. Matching errors are marked with red circles.}
    \label{fig:partial}
\end{figure}

Then we further test our method together with the baselines on point clouds undergoing three types of partiality, namely, \emph{half, hole and cut}. 
For \emph{half}, we simulate a camera in front of the point clouds and therefore capture half of the data. For \emph{hole}, we randomly choose 10 points on the surface and remove 100 nearest points around. For \emph{cut}, we randomly cut a part of the legs or arms. 
Table~\ref{tab:table5} shows the quantitative results for partial shape matching, in which we estimate point-wise maps from a partial shape to full shapes (see Fig.~\ref{fig:partial} for illustration). For \emph{hole} and \emph{cut}, the matching performance only decreases a little. As for half, though nearly half of the data are removed, our method still returns reasonable results. In particular, in Fig.~\ref{fig:partial}, we compare qualitatively our results with LIE~\cite{lie}, where we find noticeable discrepancy of the latter.  
Overall, the above results show that even trained without any ground-truth correspondence, our NIM network is capable of retrieving intrinsic information from corrupted data that are completely unseen during training.

\begin{table}[t!]
\centering
\setlength{\tabcolsep}{5mm}{
\begin{tabular}{lccc}
\hline
Method & half          & hole         & cut          \\ \hline
LIE~\cite{lie}     & 15.            & 15.         & 16.         \\
Ours   & \textbf{10.} & \textbf{7.0} & \textbf{7.2} \\ \hline
\end{tabular}}
\caption{Mean geodesic errors (×100) for partial point cloud matching.}\vspace{-2em}
\label{tab:table5}
\end{table}

\section{Conclusion, Limitations and Future Work}

To conclude, in this paper we first propose NIE, a learning based framework that embed unstructured point clouds into high-dimensional space in a way that respects intrinsic geometry of the underlying surfaces. 
Then, based on NIE, we present NIM, a weakly supervised non-rigid point cloud matching network. NIM only assumes the training point clouds to be approximately rigidly aligned, and require nothing more than geodesic distances among the training point clouds, which can even be approximated by a trained NIE. 
We demonstrate in a set of comprehensive experiments that: (1) NIE effectively learns intrinsic information and therefore allows for structured map encoding; (2) NIM enjoys decent matching performance and excellent generalization capacity; (3) Both NIE and NIM are robust to common artifacts, including noise and various partiality. 

\begin{figure}[t!]
\centering
\centerline{\includegraphics[scale=0.4]{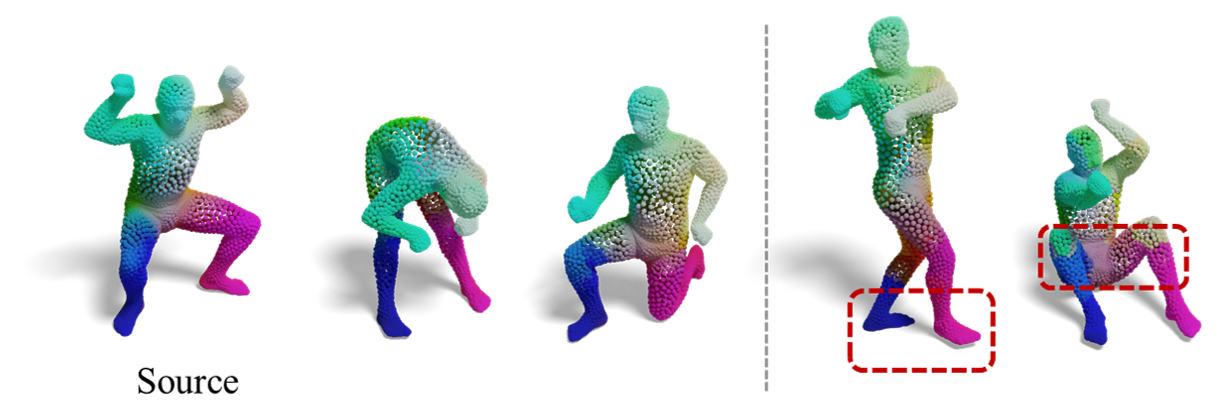}}
    \caption{Illustration of successful (left) and failure (right) cases of our method. Matching errors are marked with red rectangles.}
    \label{fig:fail}
\end{figure}

The main limitation of our framework is its sensitivity regarding extrinsic pose of point clouds. 
As shown in Fig.~\ref{fig:fail}, when shapes are reasonably aligned, our NIM can estimate high-quality maps even at the presence of significant pose differences. However, when the rigid alignment is inaccurate due to un-common poses, the estimated maps are hampered, either by severe symmetric flip(bottom middle), or erroneous intrinsic embedding (bottom right). 
It would be an interesting future work to incorporate the recent advances in $SO(3)$-invariant and -equivariant~\cite{zadeh2017tensor, deng2021vector} networks to enhance our pipeline.  
%%%%%%%%% REFERENCES
{\small
\bibliographystyle{ieee_fullname}
\bibliography{egbib}

\begin{thebibliography}{10}\itemsep=-1pt

\bibitem{anguelov2005scape}
Dragomir Anguelov, Praveen Srinivasan, Daphne Koller, Sebastian Thrun, Jim
  Rodgers, and James Davis.
\newblock Scape: shape completion and animation of people.
\newblock In {\em ACM SIGGRAPH 2005 Papers}, pages 408--416. 2005.

\bibitem{anguelov2004correlated}
Dragomir Anguelov, Praveen Srinivasan, Hoi-Cheung Pang, Daphne Koller,
  Sebastian Thrun, and James Davis.
\newblock The correlated correspondence algorithm for unsupervised registration
  of nonrigid surfaces.
\newblock {\em Advances in neural information processing systems}, 17, 2004.

\bibitem{WKS}
Mathieu Aubry, Ulrich Schlickewei, and Daniel Cremers.
\newblock The {W}ave {K}ernel {S}ignature: {A} {Q}uantum {M}echanical
  {A}pproach to {S}hape {A}nalysis.
\newblock In {\em Computer Vision Workshops (ICCV Workshops), 2011 IEEE
  International Conference on}, pages 1626--1633. IEEE, 2011.

\bibitem{pclbo}
Mikhail Belkin, Jian Sun, and Yusu Wang.
\newblock Constructing laplace operator from point clouds in ℝ d.
\newblock In {\em Proceedings of the twentieth annual ACM-SIAM symposium on
  Discrete algorithms}, pages 1031--1040. SIAM, 2009.

\bibitem{FAUST}
Federica Bogo, Javier Romero, Matthew Loper, and Michael~J. Black.
\newblock {FAUST}: Dataset and evaluation for {3D} mesh registration.
\newblock In {\em Proceedings IEEE Conf. on Computer Vision and Pattern
  Recognition (CVPR)}, Piscataway, NJ, USA, June 2014. IEEE.

\bibitem{bozic2020neural}
Aljaz Bozic, Pablo Palafox, Michael Zollh{\"o}fer, Angela Dai, Justus Thies,
  and Matthias Nie{\ss}ner.
\newblock Neural non-rigid tracking.
\newblock {\em Advances in Neural Information Processing Systems},
  33:18727--18737, 2020.

\bibitem{bozic2020deepdeform}
Aljaz Bozic, Michael Zollhofer, Christian Theobalt, and Matthias Nie{\ss}ner.
\newblock Deepdeform: Learning non-rigid rgb-d reconstruction with
  semi-supervised data.
\newblock In {\em Proceedings of the IEEE/CVF Conference on Computer Vision and
  Pattern Recognition}, pages 7002--7012, 2020.

\bibitem{bronstein2006generalized}
Alexander~M Bronstein, Michael~M Bronstein, and Ron Kimmel.
\newblock Generalized multidimensional scaling: a framework for
  isometry-invariant partial surface matching.
\newblock {\em Proceedings of the National Academy of Sciences},
  103(5):1168--1172, 2006.

\bibitem{coifman2005geometric}
Ronald~R Coifman, Stephane Lafon, Ann~B Lee, Mauro Maggioni, Boaz Nadler,
  Frederick Warner, and Steven~W Zucker.
\newblock Geometric diffusions as a tool for harmonic analysis and structure
  definition of data: Diffusion maps.
\newblock {\em Proceedings of the national academy of sciences},
  102(21):7426--7431, 2005.

\bibitem{Crane:2017:HMD}
Keenan Crane, Clarisse Weischedel, and Max Wardetzky.
\newblock The heat method for distance computation.
\newblock {\em Commun. ACM}, 60(11):90--99, Oct. 2017.

\bibitem{cui2021deep}
Yaodong Cui, Ren Chen, Wenbo Chu, Long Chen, Daxin Tian, Ying Li, and Dongpu
  Cao.
\newblock Deep learning for image and point cloud fusion in autonomous driving:
  A review.
\newblock {\em IEEE Transactions on Intelligent Transportation Systems},
  23(2):722--739, 2021.

\bibitem{deng2021vector}
Congyue Deng, Or Litany, Yueqi Duan, Adrien Poulenard, Andrea Tagliasacchi, and
  Leonidas~J Guibas.
\newblock Vector neurons: A general framework for so (3)-equivariant networks.
\newblock In {\em Proceedings of the IEEE/CVF International Conference on
  Computer Vision}, pages 12200--12209, 2021.

\bibitem{donati2020deep}
Nicolas Donati, Abhishek Sharma, and Maks Ovsjanikov.
\newblock Deep geometric functional maps: Robust feature learning for shape
  correspondence.
\newblock In {\em IEEE Conference on Computer Vision and Pattern Recognition
  (CVPR)}, June 2020.

\bibitem{eisenberger2021neuromorph}
Marvin Eisenberger, David Novotny, Gael Kerchenbaum, Patrick Labatut, Natalia
  Neverova, Daniel Cremers, and Andrea Vedaldi.
\newblock Neuromorph: Unsupervised shape interpolation and correspondence in
  one go.
\newblock In {\em Proceedings of the IEEE/CVF Conference on Computer Vision and
  Pattern Recognition}, pages 7473--7483, 2021.

\bibitem{ginzburg2020cyclic}
Dvir Ginzburg and Dan Raviv.
\newblock Cyclic functional mapping: Self-supervised correspondence between
  non-isometric deformable shapes.
\newblock In {\em European Conference on Computer Vision}, pages 36--52.
  Springer, 2020.

\bibitem{Gojcic_2019_CVPR}
Zan Gojcic, Caifa Zhou, Jan~D. Wegner, and Andreas Wieser.
\newblock The perfect match: 3d point cloud matching with smoothed densities.
\newblock In {\em The IEEE Conference on Computer Vision and Pattern
  Recognition (CVPR)}, June 2019.

\bibitem{groueix20183d}
Thibault Groueix, Matthew Fisher, Vladimir~G Kim, Bryan~C Russell, and Mathieu
  Aubry.
\newblock 3d-coded: 3d correspondences by deep deformation.
\newblock In {\em Proceedings of the European Conference on Computer Vision
  (ECCV)}, pages 230--246, 2018.

\bibitem{halimi2019unsupervised}
Oshri Halimi, Or Litany, Emanuele Rodola, Alex~M Bronstein, and Ron Kimmel.
\newblock Unsupervised learning of dense shape correspondence.
\newblock In {\em Proceedings of the IEEE/CVF Conference on Computer Vision and
  Pattern Recognition}, pages 4370--4379, 2019.

\bibitem{huang2014functional}
Qixing Huang, Fan Wang, and Leonidas Guibas.
\newblock Functional map networks for analyzing and exploring large shape
  collections.
\newblock {\em ACM Transactions on Graphics (TOG)}, 33(4):1--11, 2014.

\bibitem{huang2008non}
Qi-Xing Huang, Bart Adams, Martin Wicke, and Leonidas~J Guibas.
\newblock Non-rigid registration under isometric deformations.
\newblock In {\em Computer Graphics Forum}, volume~27, pages 1449--1457. Wiley
  Online Library, 2008.

\bibitem{huang2017adjoint}
Ruqi Huang and Maks Ovsjanikov.
\newblock Adjoint map representation for shape analysis and matching.
\newblock In {\em Computer Graphics Forum}, volume~36, pages 151--163. Wiley
  Online Library, 2017.

\bibitem{jin2019fast}
Young-Hoon Jin and Won-Hyung Lee.
\newblock Fast cylinder shape matching using random sample consensus in large
  scale point cloud.
\newblock {\em Applied Sciences}, 9(5):974, 2019.

\bibitem{geod}
R. Kimmel and J. Sethian.
\newblock Fast marching methods on triangulated domains.
\newblock In {\em PNAS}, 1998.

\bibitem{kovnatsky2013coupled}
Artiom Kovnatsky, Michael~M Bronstein, Alexander~M Bronstein, Klaus Glashoff,
  and Ron Kimmel.
\newblock Coupled quasi-harmonic bases.
\newblock In {\em Computer Graphics Forum}, volume~32, pages 439--448. Wiley
  Online Library, 2013.

\bibitem{li2018articulatedfusion}
Chao Li, Zheheng Zhao, and Xiaohu Guo.
\newblock Articulatedfusion: Real-time reconstruction of motion, geometry and
  segmentation using a single depth camera.
\newblock In {\em Proceedings of the European Conference on Computer Vision
  (ECCV)}, pages 317--332, 2018.

\bibitem{li2018robust}
Kun Li, Jingyu Yang, Yu-Kun Lai, and Daoliang Guo.
\newblock Robust non-rigid registration with reweighted position and
  transformation sparsity.
\newblock {\em IEEE transactions on visualization and computer graphics},
  25(6):2255--2269, 2018.

\bibitem{li2020learning}
Yang Li, Aljaz Bozic, Tianwei Zhang, Yanli Ji, Tatsuya Harada, and Matthias
  Nie{\ss}ner.
\newblock Learning to optimize non-rigid tracking.
\newblock In {\em Proceedings of the IEEE/CVF Conference on Computer Vision and
  Pattern Recognition}, pages 4910--4918, 2020.

\bibitem{lipman2010biharmonic}
Yaron Lipman, Raif~M Rustamov, and Thomas~A Funkhouser.
\newblock Biharmonic distance.
\newblock {\em ACM Transactions on Graphics (TOG)}, 29(3):1--11, 2010.

\bibitem{litany2017deep}
Or Litany, Tal Remez, Emanuele Rodol{\`a}, Alex Bronstein, and Michael
  Bronstein.
\newblock Deep functional maps: Structured prediction for dense shape
  correspondence.
\newblock In {\em Proceedings of the IEEE International Conference on Computer
  Vision}, pages 5659--5667, 2017.

\bibitem{lie}
Riccardo Marin, Marie-Julie Rakotosaona, Simone Melzi, and Maks Ovsjanikov.
\newblock Correspondence learning via linearly-invariant embedding.
\newblock {\em Advances in Neural Information Processing Systems},
  33:1608--1620, 2020.

\bibitem{marin2021spectral}
Riccardo Marin, Arianna Rampini, Umberto Castellani, Emanuele Rodol{\`a}, Maks
  Ovsjanikov, and Simone Melzi.
\newblock Spectral shape recovery and analysis via data-driven connections.
\newblock {\em International journal of computer vision}, 129(10):2745--2760,
  2021.

\bibitem{moschella2021spectral}
Luca Moschella, Simone Melzi, Luca Cosmo, Filippo Maggioli, Or Litany, Maks
  Ovsjanikov, Leonidas Guibas, and Emanuele Rodol{\`a}.
\newblock Spectral unions of partial deformable 3d shapes.
\newblock {\em arXiv preprint arXiv:2104.00514}, 2021.

\bibitem{nogneng2017informative}
Dorian Nogneng and Maks Ovsjanikov.
\newblock Informative descriptor preservation via commutativity for shape
  matching.
\newblock In {\em Computer Graphics Forum}, volume~36, pages 259--267. Wiley
  Online Library, 2017.

\bibitem{ovsjanikov2012functional}
Maks Ovsjanikov, Mirela Ben-Chen, Justin Solomon, Adrian Butscher, and Leonidas
  Guibas.
\newblock Functional maps: a flexible representation of maps between shapes.
\newblock {\em ACM Transactions on Graphics (TOG)}, 31(4):30:1--30:11, 2012.

\bibitem{pai2021fast}
Gautam Pai, Jing Ren, Simone Melzi, Peter Wonka, and Maks Ovsjanikov.
\newblock Fast sinkhorn filters: Using matrix scaling for non-rigid shape
  correspondence with functional maps.
\newblock In {\em Proceedings of the IEEE/CVF Conference on Computer Vision and
  Pattern Recognition}, pages 384--393, 2021.

\bibitem{paravati2016point}
Gianluca Paravati, Fabrizio Lamberti, Valentina Gatteschi, Claudio Demartini,
  and Paolo Montuschi.
\newblock Point cloud-based automatic assessment of 3d computer animation
  courseworks.
\newblock {\em IEEE Transactions on Learning Technologies}, 10(4):532--543,
  2016.

\bibitem{paszke2019pytorch}
Adam Paszke, Sam Gross, Francisco Massa, Adam Lerer, James Bradbury, Gregory
  Chanan, Trevor Killeen, Zeming Lin, Natalia Gimelshein, Luca Antiga, et~al.
\newblock Pytorch: An imperative style, high-performance deep learning library.
\newblock {\em Advances in neural information processing systems}, 32, 2019.

\bibitem{cotangent}
Ulrich Pinkall and Konrad Polthier.
\newblock Computing discrete minimal surfaces and their conjugates.
\newblock In {\em Experimental Mathematics}, 1993.

\bibitem{ren2018continuous}
Jing Ren, Adrien Poulenard, Peter Wonka, and Maks Ovsjanikov.
\newblock Continuous and orientation-preserving correspondences via functional
  maps.
\newblock {\em ACM Transactions on Graphics (TOG)}, 37(6):1--16, 2018.

\bibitem{reuter2006laplace}
Martin Reuter, Franz-Erich Wolter, and Niklas Peinecke.
\newblock Laplace--beltrami spectra as ‘shape-dna’of surfaces and solids.
\newblock {\em Computer-Aided Design}, 38(4):342--366, 2006.

\bibitem{roufosse2019unsupervised}
Jean-Michel Roufosse, Abhishek Sharma, and Maks Ovsjanikov.
\newblock Unsupervised deep learning for structured shape matching.
\newblock In {\em Proceedings of the IEEE International Conference on Computer
  Vision}, pages 1617--1627, 2019.

\bibitem{lbo}
Raif~M Rustamov et~al.
\newblock Laplace-beltrami eigenfunctions for deformation invariant shape
  representation.
\newblock In {\em Symposium on geometry processing}, volume 257, pages
  225--233, 2007.

\bibitem{sharma2020weakly}
Abhishek Sharma and Maks Ovsjanikov.
\newblock Weakly supervised deep functional maps for shape matching.
\newblock {\em Advances in Neural Information Processing Systems},
  33:19264--19275, 2020.

\bibitem{sharp2020laplacian}
Nicholas Sharp and Keenan Crane.
\newblock A laplacian for nonmanifold triangle meshes.
\newblock In {\em Computer Graphics Forum}, volume~39, pages 69--80. Wiley
  Online Library, 2020.

\bibitem{sun2009concise}
Jian Sun, Maks Ovsjanikov, and Leonidas Guibas.
\newblock A concise and provably informative multi-scale signature based on
  heat diffusion.
\newblock {\em Computer Graphics Forum}, 28(5):1383--1392, 2009.

\bibitem{SANCHEZ2020}
Carlos Sánchez-Belenguer, Simone Ceriani, Pierluigi Taddei, Erik Wolfart, and
  Vítor Sequeira.
\newblock Global matching of point clouds for scan registration and loop
  detection.
\newblock {\em Robotics and Autonomous Systems}, 123:103324, 2020.

\bibitem{tombari2010unique}
Federico Tombari, Samuele Salti, and Luigi~Di Stefano.
\newblock Unique signatures of histograms for local surface description.
\newblock In {\em European conference on computer vision}, pages 356--369.
  Springer, 2010.

\bibitem{torgerson1952multidimensional}
Warren~S Torgerson.
\newblock Multidimensional scaling: I. theory and method.
\newblock {\em Psychometrika}, 17(4):401--419, 1952.

\bibitem{varol2017learning}
Gul Varol, Javier Romero, Xavier Martin, Naureen Mahmood, Michael~J Black, Ivan
  Laptev, and Cordelia Schmid.
\newblock Learning from synthetic humans.
\newblock In {\em Proceedings of the IEEE conference on computer vision and
  pattern recognition}, pages 109--117, 2017.

\bibitem{wang2019dynamic}
Yue Wang, Yongbin Sun, Ziwei Liu, Sanjay~E Sarma, Michael~M Bronstein, and
  Justin~M Solomon.
\newblock Dynamic graph cnn for learning on point clouds.
\newblock {\em Acm Transactions On Graphics (tog)}, 38(5):1--12, 2019.

\bibitem{wu2019global}
Zhenchao Wu, Kun Li, Yu-Kun Lai, and Jingyu Yang.
\newblock Global as-conformal-as-possible non-rigid registration of multi-view
  scans.
\newblock In {\em 2019 IEEE International Conference on Multimedia and Expo
  (ICME)}, pages 308--313. IEEE, 2019.

\bibitem{xia2021geodesicembedding}
Qianwei Xia, Juyong Zhang, Zheng Fang, Jin Li, Mingyue Zhang, Bailin Deng, and
  Ying He.
\newblock Geodesicembedding (ge): a high-dimensional embedding approach for
  fast geodesic distance queries.
\newblock {\em IEEE Transactions on Visualization and Computer Graphics}, 2021.

\bibitem{xu2019unstructuredfusion}
Lan Xu, Zhuo Su, Lei Han, Tao Yu, Yebin Liu, and Lu Fang.
\newblock Unstructuredfusion: Realtime 4d geometry and texture reconstruction
  using commercial rgbd cameras.
\newblock {\em IEEE transactions on pattern analysis and machine intelligence},
  42(10):2508--2522, 2019.

\bibitem{yang2019global}
Jingyu Yang, Daoliang Guo, Kun Li, Zhenchao Wu, and Yu-Kun Lai.
\newblock Global 3d non-rigid registration of deformable objects using a single
  rgb-d camera.
\newblock {\em IEEE Transactions on Image Processing}, 28(10):4746--4761, 2019.

\bibitem{yue2018lidar}
Xiangyu Yue, Bichen Wu, Sanjit~A Seshia, Kurt Keutzer, and Alberto~L
  Sangiovanni-Vincentelli.
\newblock A lidar point cloud generator: from a virtual world to autonomous
  driving.
\newblock In {\em Proceedings of the 2018 ACM on International Conference on
  Multimedia Retrieval}, pages 458--464, 2018.

\bibitem{zadeh2017tensor}
Amir Zadeh, Minghai Chen, Soujanya Poria, Erik Cambria, and Louis-Philippe
  Morency.
\newblock Tensor fusion network for multimodal sentiment analysis.
\newblock {\em arXiv preprint arXiv:1707.07250}, 2017.

\bibitem{zeng2021corrnet3d}
Yiming Zeng, Yue Qian, Zhiyu Zhu, Junhui Hou, Hui Yuan, and Ying He.
\newblock Corrnet3d: Unsupervised end-to-end learning of dense correspondence
  for 3d point clouds.
\newblock In {\em Proceedings of the IEEE/CVF Conference on Computer Vision and
  Pattern Recognition}, pages 6052--6061, 2021.

\end{thebibliography}
}

\clearpage \appendix
\label{sec:appendix}
\section{Implementation Details}
\begin{figure*}[!hb]
\centerline{\includegraphics[scale=0.6]{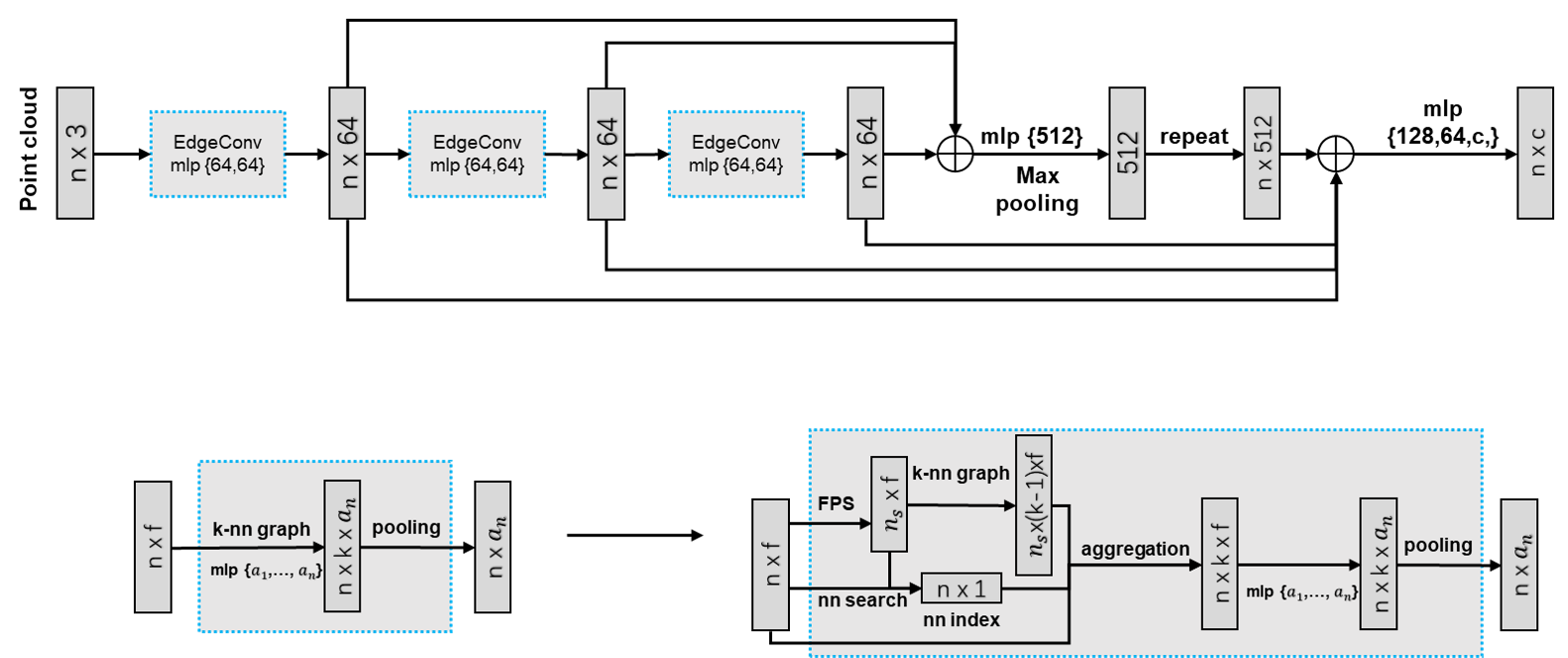}}
    \caption{The top row depicts the network architecture of DGCNN~\cite{wang2019dynamic}. In the bottom row, we specify our modification on top of the EdgeConv blocks. Taking input dimension $f = 3$ as an example: Given a point cloud $X$, we subsample $X_s$ by FPS sampling. Then we conduct nearest neighbor search of $X$ on $X_s$ and $k$-NN search(exclude itself) on $X_s$. After that, for a query point $v_q$ in $X$, we find its nearest neighbor $v_p$ within $X_s$ and assign the $k$-NN of $v_p$ within $X_s$ to that of $v_q$. Finally, we concatenate the $k$-NN of $v_p$ and $v_q$ as the aggregation feature to next procedure. }\vspace{-1em}
    \label{fig:dgcnn}
\end{figure*}

\vspace{-2em}
\noindent\paragraph*{Basis and Descriptor Setting} For a fair comparison to {LIE}\cite{lie}, we set the dimension of the output basis of NIE to be $20$. Similarly, for the matching network, NIM, we set the dimension of learned features (descriptors) to be $40$. Considering NIE as a key component, we conduct an ablation study on the dimension of basis in Table \ref{tab:table7}. As dim increases, OPT is improved while GeoError is worse (probably due to over-fitting). Setting it to $20$ is a reasonable trade-off. In Fig.\ref{fig:dgcnn}, we provide the detailed network architecture and parameters.
\begin{table}[h]
\centering
\begin{tabular}{lccc}
\hline
Dimension & 10  & 20  & 30\\ \hline
  OPT            & 4.7      &3.1          &    2.9          \\
    Geo Error        &   7.1     &    9.5     &  11.3                   \\ \hline
\end{tabular}
\caption{Ablation study of basis dimension.}\vspace{-1em}
\label{tab:table7}
\end{table}
\noindent\paragraph*{Down-sampling Scheme on the Modified DGCNN}
Recall that in Section 4 or the main submission, we propose a modified version of DGCNN~\cite{wang2019dynamic}, which leverages point cloud down-sampling for alleviating sampling density bias. We denote by $n_s$ the size of the sub-sampled point obtained from furthest point sampling. The empirical test validates that $n_s=3000$ achieves a good balance between efficiency and accuracy for all the datasets considered in our paper.

\noindent\paragraph*{Additional details on experiments}
We set $ \lambda_{1} = 1$, $ \lambda_{2} = 1$ and $ \lambda_{3} = 0.5$ in Eqn.~\ref{eqn:total} for our experiments.  We run a line search of $\alpha$ in Eqn.~\ref{eqn:alpha} on small-scale dataset and fix it to be $30$ for \emph{all} experiments.

\section{Geodesic Approximation for Partial Point Clouds}
\begin{figure}[ht]
\centerline{\includegraphics[scale=0.27]{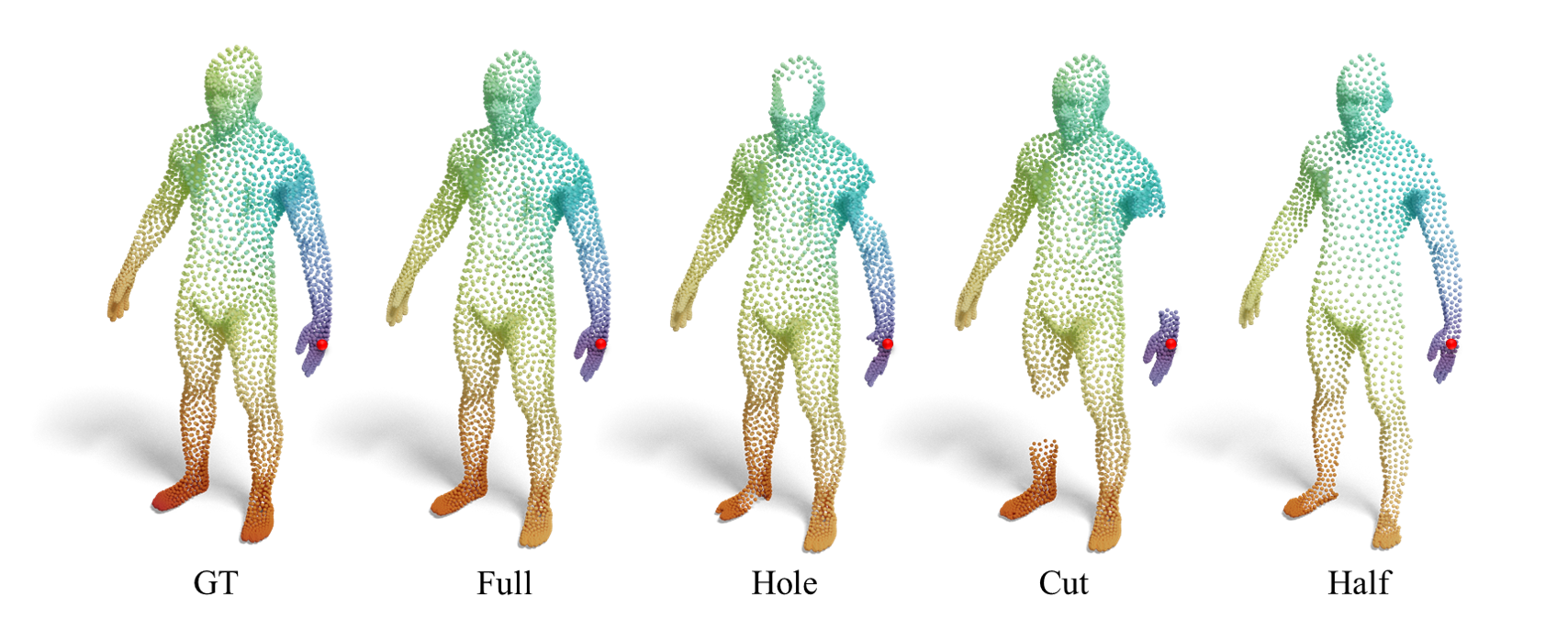}}
    \caption{Geodesic distances approximated by NIE on partial point clouds. In each input, we set the source point as the red dot on the left hand, and visualize the geodesic distance from all the other point to it. The color ranges from blue (small distance) to red (large distance). }\vspace{-1em}
    \label{fig:geo}
\end{figure}

Unlike the mesh~\cite{geod} or Laplacian~\cite{Crane:2017:HMD} based approaches, our approach can robustly approximate geodesic distances on disconnected shapes (see, e.g., the $\emph{hole}$ and the $\emph{cut}$ in Fig.~\ref{fig:geo}). Table~\ref{tab:table6} shows that NIE maintains a reasonable geodesic error when partial point clouds (generated with FAUST\_r dataset) are given. Fig.\ref{fig:geo} shows the qualitative examples of geodesic distance, where the source points are all set in the left hand.

\begin{table}[t!]
\centering
\setlength{\tabcolsep}{5mm}{
\begin{tabular}{lcccc}
\hline
Method & full & half          & hole         & cut          \\ \hline
Ours  & 9.5   & 9.8 & 10. & 13.0 \\ \hline
\end{tabular}}
\caption{Relative geodesic errors (×100) for full and partial point clouds.}\vspace{-2em}
\label{tab:table6}
\end{table}

\end{document}